\pdfoutput=1

\documentclass[11pt]{article}

\usepackage{EMNLP2023}

\usepackage{times}
\usepackage{latexsym}
\usepackage{amsmath,amsthm,amssymb,amsfonts,amscd,mathrsfs}
\usepackage{graphicx}
\usepackage{caption}
\usepackage{subfig}

\usepackage[T1]{fontenc}

\def\R{\mathbb R}

\usepackage[utf8]{inputenc}

\usepackage{microtype}

\usepackage{inconsolata}

\title{MLLM-Microscope: Unlocking Hidden Structure Within Multimodal Large Language Models}

\author{Ravil Mussabayev \\
  Satbayev University \\
  \texttt{r.mussabayev@satbayev.university} \\\And
  Rustam Mussabayev \\
  Satbayev University \\
  \texttt{ru.mussabayev@satbayev.university} \\}

\begin{document}
\maketitle
\begin{abstract}

This work presents MLLM-Microscope, a novel system designed for analyzing the hidden representations within Multimodal Large Language Models (MLLMs). Our system evaluates the linearity, intrinsic dimension, and anisotropy of multimodal token embeddings across transformer layers. Utilizing the ScienceQA dataset, we evaluate two state-of-the-art MLLMs, LLaVA-NeXT and OmniFusion. We find that both the main and residual streams for tokens of both modalities exhibit highly linear behaviors across transformer layers. However, LLaVA-NeXT's image tokens reveal a slight decline in linearity, whereas OmniFusion's remain consistent. Image token dimensions in OmniFusion remain consistently higher across layers compared to LLaVA-NeXT. Also, the OmniFusion's anisotropy is observed to stay consistently low throughout the layers. These findings suggest that the inner workings of MLLMs highly depend on the nature of modality fusion performed before passing the token sequence into LLM. This and other new potential insights obtainable from our system are surely capable of enhancing our understanding of the inner workings of MLLMs, informing future model design and optimization.
\end{abstract}

\section{Introduction}

Recently, large language models (LLMs) have provided remarkable emergent capabilities over their smaller counterparts, including instruction following, In-Context Learning (ICL), and Chain of Thought (CoT) \cite{Zhao2023-llmsurvey}. Although LLMs have shown outstanding zero or few-shot performance on most downstream text-related tasks, intrinsically they are unable to understand other modalities, such as vision, sound, video, etc.

CLIP \cite{Radford2021-clip} was an early work that explored tying different modalities together in a single embedding space through contrastive training. Also, there have been numerous research works aimed at achieving modality fusion within LLMs, such as Flamingo, InstructBLIP, LLaVA-NeXT, and many others \cite{Yin2024-mllmsurvey}. The large variety of emerging multimodal architectures necessitates advanced tools for their comparative analysis \cite{Zhang2024-mllmssurvey}. An interesting direction for investigation by our proposed system is the distinction between various modality fusion techniques in terms of their impact on the evolution of multimodal tokens through the transformer layers of an underlying LLM. For instance, LLaVA \cite{Liu2023-llava} is an example of shallow fusion, while OmniFusion \cite{Goncharova2024-omnifusion} employs a deeper fusion technique due to the use of a transformer-based adapter.

Nowadays, most LLMs are based on transformer architectures, and little is known about their inner workings. More detailed insights in this direction will prove highly valuable for optimizing the performance of transformer architectures and, consequently, LLMs. The papers by \cite{Nostalgebraist2020-logitlens} and \cite{Belrose2023-tunedlens} introduced the techniques called ``logit lens'' and ``tuned lens'' for understanding how model predictions are refined layer by layer relative to the final layer output. However, these works fell short of analyzing the inherent linearity of transformations between layers. The authors of \cite{Elhage2021-transcircuits} discovered a hidden linear structure present in small transformer models via a novel mathematical framework. Then, (Razzhigaev et al., 2024) profoundly expanded this line of research by observing that the transformations between consecutive layers in transformer decoders are almost perfectly linear with Procrustes similarity score of $0.99$.

The goal of the present work is to develop a convenient tool for the inspection of Multimodal Large Language Models (MLLMs). The tool allows to examine how the hidden representations of textual and visual tokens alter across transformer layers in terms of linearity (measured by the Procrustes similarity score), inner dimensionality, and anisotropy. Additionally, our system provides an illuminating visualization of each transformation step. We hope that the presented tool will be useful for researchers and practitioners in gaining insights into the internal mechanisms of multimodal LLMs, driving improvements in the design or training/inference procedures of these models.

Our main contributions are the following:
\begin{enumerate}
\item To the best of our knowledge, we are the first to propose an efficient and effective system for analyzing the evolution of multimodal token representations across transformer layers of MLLMs;
\item The system is tested on the two recent state-of-the-art MLLMs: LLaVA-NeXT and OmniFusion;
\item A comparative analysis between these models is conducted based on the system's output, revealing new emergent characteristics of multimodal tokens in MLLMs and insights about their hidden structure. Our findings are capable of driving and guiding future, more optimal model designs. For instance, they suggest that using a cosine-similarity-based regularization, aimed
at reducing layer linearity \cite{Razzhigaev2024-secretlylinear} might be highly relevant for the MLLMs with deeper modality fusion modules.
\end{enumerate}

The rest of the article is structured as follows.

\section{Methodology}

Our tool focuses on understanding the hidden representations in the transformer of an MLLM from multiple sides. Specifically, it examines the degree of linearity and smoothness of transformations between each pair of sequential layers $L_i$ and $L_{i + 1}$, as well as estimates the intrinsic dimensions and anisotropy scores of representations at every layer.

\subsection{Linearity score}

Let $T_i, T_{i + 1} \in \R^{n_t \times d}$ and $V_i, V_{i + 1} \in \R^{n_v \times d}$ be the centered sets of embeddings of textual and visual (image) tokens at two consecutive layers, where $n_t$ and $n_v$ represent the number of textual and visual tokens in the input sequence, respectively.

To measure the linear dependence between two consecutive sets of embeddings, we used a generalized Procrustes similarity metric \cite{Gower1975-procrustes}. We chose this approach for its robustness in evaluating the linearity of embeddings, particularly given the varying scales across transformer layers. Unlike the $L_2$ norm, which is not scale-invariant, Procrustes normalization provides a bounded metric ranging from 0 to 1 \cite{Razzhigaev2024-secretlylinear}.

Using tildes to denote matrices normalized by the Frobenius norm, the linearity scores are computed as follows:
\begin{align}
  \textrm{PS}_t(i) &:= 1 - \min_{A \in \R^{d \times d}} \| \tilde{T_i}A - \tilde{T}_{i+1} \|_2^2 \label{eq:ps_t}, \\
  \textrm{PS}_v(i) &:= 1 - \min_{A \in \R^{d \times d}} \| \tilde{V_i}A - \tilde{V}_{i+1} \|_2^2 \label{eq:ps_i}.
\end{align}

These formulas closely resemble the Procrustes similarity, but instead of minimizing over orthogonal transformations, all linear transformations are used to minimize the squared errors \cite{Razzhigaev2024-secretlylinear}.

The centered versions of the linearity scores \eqref{eq:ps_t} and \eqref{eq:ps_i} can be defined as follows:
\begin{align}
  \textrm{PSC}_t(i) &:= 1 - \min_{A \in \R^{d \times d}} \| \tilde{T_i}A - (\tilde{T}_{i+1} - \tilde{T_i}) \|_2^2 \label{eq:psc_t}, \\
  \textrm{PSC}_v(i) &:= 1 - \min_{A \in \R^{d \times d}} \| \tilde{V_i}A - (\tilde{V}_{i+1} - \tilde{V_i}) \|_2^2 \label{eq:psc_i}.
\end{align}

These scores allow to estimate the linearity in the main flow of the transformer excluding the residuals.

\subsection{Anisotropy}

Anisotropy, essentially representing the non-uniformity of a distribution in space, provides a lens, through which we can study orientation and
concentration of embeddings \cite{Ethayarajh2019-contextual,Bis2021-toomuch}.

To compute anisotropy, we employ the singular value decomposition (SVD). Let $X \in \{V_i, T_i\}$ be a centered matrix of either textual or visual embeddings extracted from the $i$-th layer, and $\sigma_1, \dots, \sigma_k$ be its singular values. Then, the anisotropy score of $X$ is defined as:
$$
\textrm{A}(X) = \frac{\sigma_1^2}{\sum_{i=1}^{k} \sigma_i^2}.
$$
The singular values can be obtained by computing the eigenvalues $\sigma_1^2, \dots, \sigma_k^2$ of the covariance matrix:
$$
C = \frac{X^T X}{n\_\mathrm{samples}(X) - 1}.
$$

\subsection{Intrinsic dimension}

The intrinsic dimension offers a measure of the effective data dimensionality, highlighting the essence of information that is captured by the embeddings. Together, these metrics can serve as pivotal tools to probe into the black-box nature of transformers.

To determine the intrinsic dimension of a set of embeddings, we employ the methodology proposed by \cite{Facco2018-intdim} and \cite{Razzhigaev2024-secretlylinear}. This technique examines how the volume of an $n$-dimensional sphere, which represents the number of embeddings, scales with the dimension $d$.

For each data point in our embeddings, we calculate the distances $r_1$ and $r_2$ to their two nearest neighboring points, resulting in a set of pairs $\{(r_1, r_2)\}$. Using this set, we can estimate the intrinsic dimension $d$. Specifically, we define:
$$
\mu_i = \frac{r_2}{r_1},
$$
for each point $i$.

The cumulative distribution function (CDF) of $\{\mu_i\}$ is given by:
$$
F(\mu) = (1 - \mu^{-d}) \mathbf{1}_{[1,+\infty)}(\mu).
$$
This form of $F$ is derived from the results and proofs provided by the authors of the cited works. From the CDF, we infer:
$$
\frac{\log(1 - F(\mu))}{\log(\mu)} = d.
$$

To estimate $d$, we apply linear regression $y = kx$ on the plane $(x, y)$, with:
$$
x_i = \log(\mu_i) \quad \text{and} \quad y_i = 1 - F_{\text{emp}}(\mu_i),
$$
where $F_{\text{emp}}$ denotes the empirical CDF for $\{\mu_i\}$.

\section{Dataset}

The \textit{ScienceQA} dataset \cite{Lu2022-scienceqa} was used for the experiments. ScienceQA consists of 21,208 multimodal multiple-choice science questions sourced from elementary and high school curricula. It features diverse science topics and annotations of answers with corresponding lectures and explanations, making it the first large-scale multimodal dataset to do so.

The dataset includes:
\begin{itemize}
    \item 10,332 questions with an image context (48.7\%);
    \item 10,220 questions with a text context (48.2\%);
    \item 6,532 questions with both image and text contexts (30.8\%);
    \item 83.9\% of questions annotated with grounded lectures;
    \item 90.5\% of questions annotated with detailed explanations.
\end{itemize}

ScienceQA covers a wide range of domains, categorized by subject (natural science, language science, social science), topic (e.g., Biology, Physics, Chemistry), and skill (e.g., classify fruits and vegetables as plant parts). This rich domain diversity enhances the evaluation of multi-hop reasoning ability and interpretability of AI systems. The dataset supports tasks such as multi-modal multiple-choice question answering in English.

\section{Models}

We have selected two competitive state-of-the-art MLLMs for the experimental analysis, LLaVA-NeXT 1.6 and OmniFusion, both of which leverage the Mistral-7b LLM in half precision as their backbone.

LLaVA (Large Language and Vision Assistant) \cite{Liu2023-llava} is an end-to-end trained large multimodal model, which connects a vision encoder and an LLM for impressive results on natural instruction-following and visual reasoning capabilities. The pre-trained CLIP visual encoder provides the visual features $Z_0 = \textrm{CLIP}(input\_image)$, and a simple linear layer (trainable projection matrix $W$) is used to fuse them into the LLM's word embedding space: $V_0 = W \cdot Z_0$. Thus, LLaVA is an prominent example of the shallow modality fusion technique. This simple projection scheme is very lightweight and time-efficient. The training of LLaVA was conducted on multimodal multi-turn conversation instruction-following sequences and consisted of two stages: pre-training and end-to-end fine-tuning.

In pre-training, we keep both the visual encoder and LLM weights frozen, and maximize the auto-regressive training objective with trainable parameters $W$ (the projection matrix) only. This stage can be understood as training a compatible visual tokenizer for the frozen LLM. In fine-tuning, we always keep the visual encoder weights frozen, and continue to update both the pre-trained weights of the projection layer and LLM in LLaVA.

The subsequent work \cite{Liu2023-improvedllava} discovered that the original LLaVA model's multimodal understanding capabilities could be significantly enhanced with a few simple improvements: swapping the vision encoder to high-resolution CLIP-ViT-L-336px, incorporating a two-layer MLP cross-modal connector, and additional training on academic-task-related data such as VQA. The LLaVA-NeXT model \cite{Liu2024-llavanext} presented several further improvements, including a fourfold increase in input image resolution, enhanced visual reasoning and OCR capabilities with a better visual instruction tuning data mixture, and improved visual conversation across various scenarios.

The OmniFusion model \cite{Goncharova2024-omnifusion} is based on a strong pretrained LLM (Mistral-7B) and powerful adapters for visual modality. OmniFusion also features special tokens of new modalities and a multi-stage, end-to-end instructional learning with the gradual unfreezing of the LLM. The training data consisted of a combination of open-sourced synthetic instruction-following data and high-quality image-language datasets, curated for various downstream tasks. Early experiments revealed that OmniFusion matched multimodal GPT-4V in chat capabilities on unseen images and instructions, achieving comparable benchmark results to larger models in various use cases.

\section{Experiments}

\subsection{Hardware}

The experiments were conducted on a high-performance computational system with the following hardware specifications: 2x AMD EPYC 7663 56-Core Processor (100 CPUs), 2x NVIDIA A100 80GB GPUs, 1.5 TiB RAM, and 3 TiB storage. This configuration provides substantial computational power, enabling efficient execution of large-scale machine learning experiments.

\subsection{Software}

The software environment is set up on a Linux-based system equipped with Python 3.10.14 and the following key libraries installed for running LLMs: \texttt{torch} 2.4.0, \texttt{transformers} 4.43.3, \texttt{datasets} 2.20.0, \texttt{numpy} 1.26.4, and \texttt{scikit-learn} 1.5.1. This software setup ensures a reliable environment for inference of large-scale language models.

\subsection{Implementation details}

Our system's package is written in Python and bases on the LLM analysis framework developed in \cite{Razzhigaev2024-secretlylinear}.

Using the Huggingface's \texttt{Transformers} API, our code precisely identifies the positions of text and image tokens in every input sequence according to the peculiar rules defined by each model. For every layer, the embeddings from each input sequence are collected token-wise into two separate tensors holding the two modalities: text and image.

To avoid evaluating the models on the data they have already seen, the experimental tests were performed on the test split of the ScienceQA dataset. For instance, LLaVA's original paper mentioned training their model on the ScienceQA dataset.

\subsection{Downloadable package}

The downloadable package with the full source code and a demo Jupyter notebook is available at GitHub: \href{https://github.com/rmusab/mllm-microscope}{https://github.com/rmusab/mllm-microscope}

\section{Discussion of Results}

\begin{figure*}[htbp]
    \centering
    \subfloat[Procrustes similarity for textual tokens\label{fig:proc_sim_txt}]{
        \includegraphics[width=0.3\textwidth]{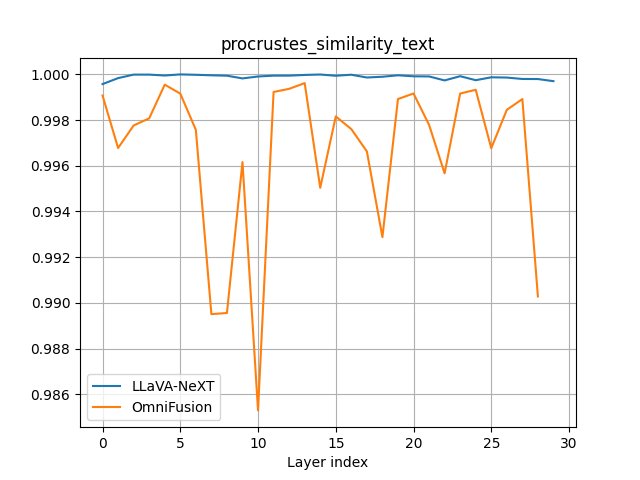}
    }
    \hfill
    \subfloat[Procrustes similarity for image tokens\label{fig:proc_sim_img}]{
        \includegraphics[width=0.3\textwidth]{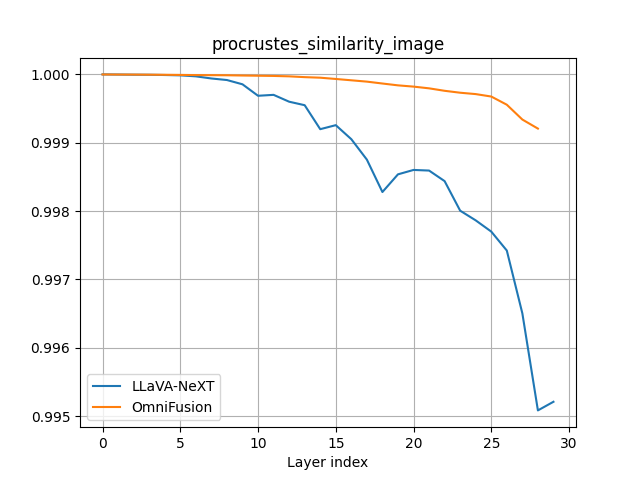}
    }
    \hfill
    \subfloat[Centered Procrustes similarity for textual tokens\label{fig:proc_sim_cen_txt}]{
        \includegraphics[width=0.3\textwidth]{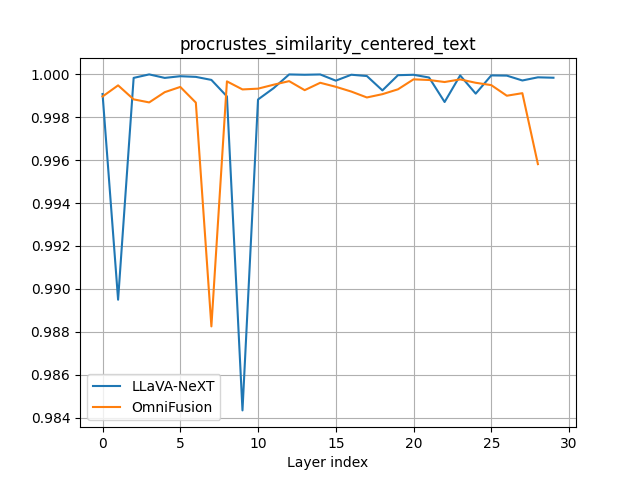}
    }
    \\
    \subfloat[Centered Procrustes similarity for image tokens\label{fig:proc_sim_cen_img}]{
        \includegraphics[width=0.3\textwidth]{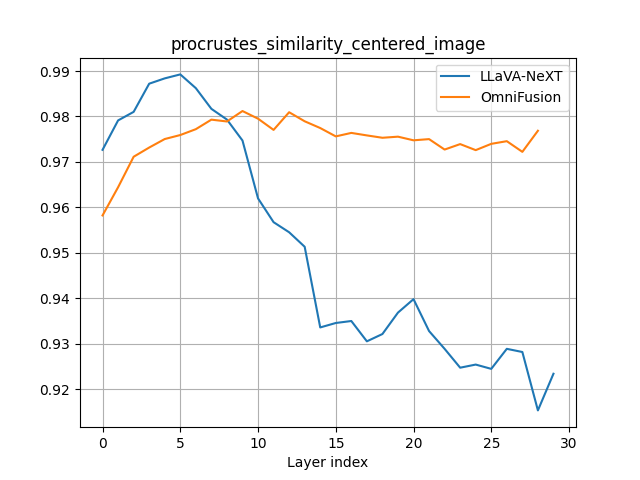}
    }
    \hfill
    \subfloat[Intrinsic dimension for textual tokens\label{fig:id_txt}]{
        \includegraphics[width=0.3\textwidth]{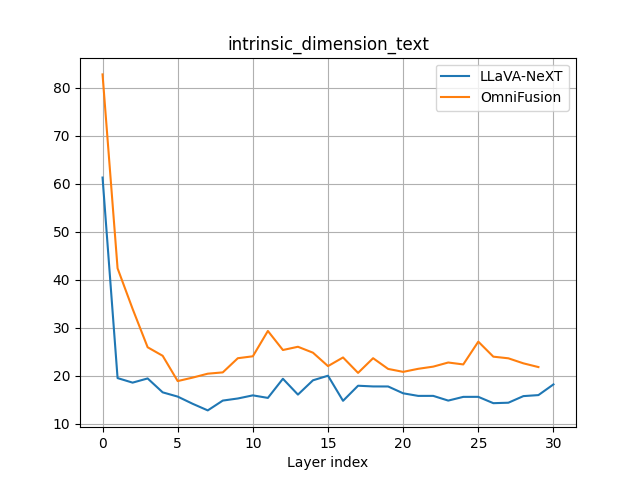}
    }
    \hfill
    \subfloat[Intrinsic dimension for image tokens\label{fig:id_img}]{
        \includegraphics[width=0.3\textwidth]{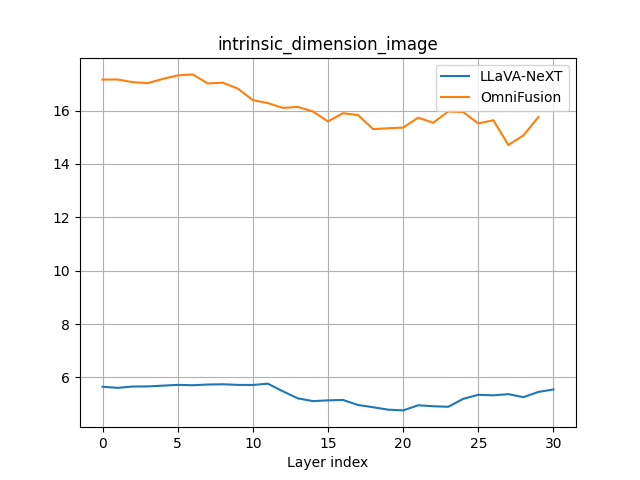}
    }
    \\
    \subfloat[Intrinsic dimension for the joint text-image token distribution\label{fig:id_joint}]{
        \includegraphics[width=0.3\textwidth]{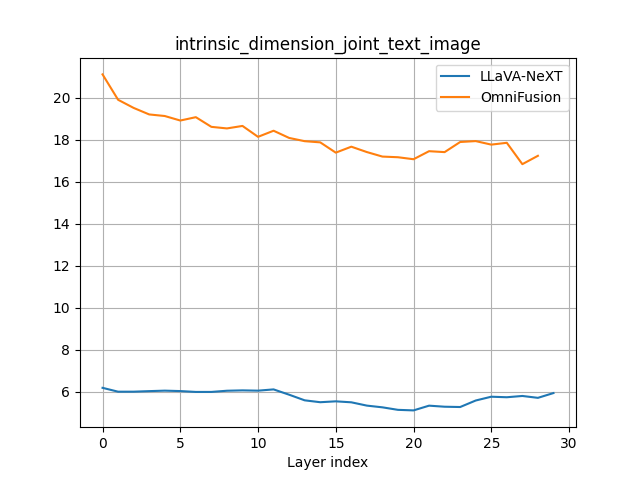}
    }
    \hfill
    \subfloat[Anisotropy for textual tokens\label{fig:aniso_txt}]{
        \includegraphics[width=0.3\textwidth]{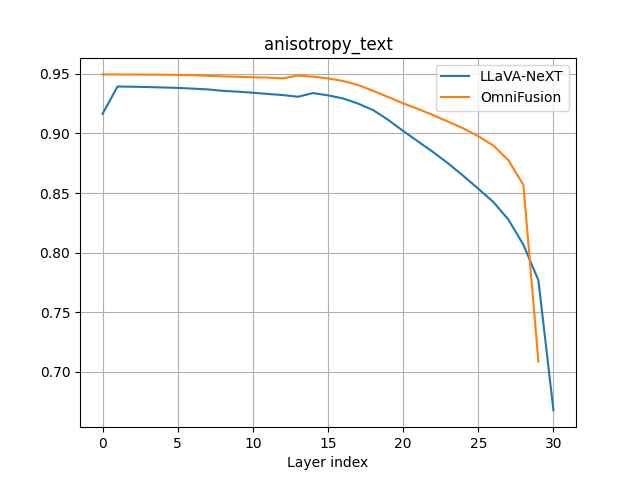}
    }
    \hfill
    \subfloat[Anisotropy for image tokens\label{fig:aniso_img}]{
        \includegraphics[width=0.3\textwidth]{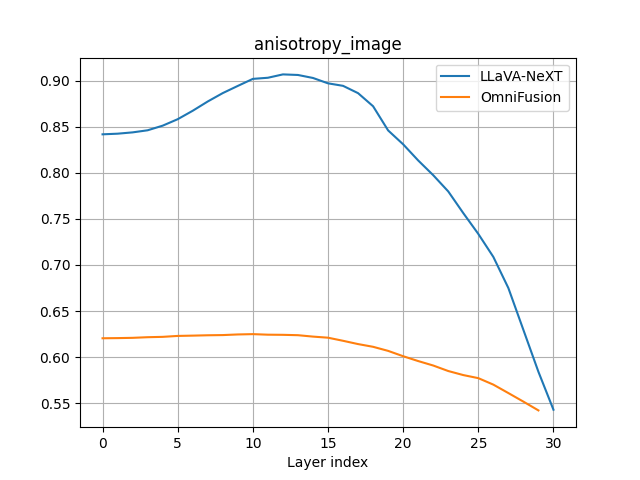}
    }
    \caption{Plots illustrating the evolution of textual and visual tokens across intermediate transformer layers for the examined LLaVA-NeXT and OmniFusion models using various scores: uncentered and centered Procrustes similarity, intrinsic dimensions, and anisotropies of intermediate representations.}
    \label{fig:all_figs}
\end{figure*}

The results of running the analysis on 16 random examples from the test split are presented in Figures \ref{fig:proc_sim_txt}-\ref{fig:aniso_img}. In total, 1563 textual and 21964 visual token embeddings of size 4096 were analyzed for both models across 32 transformer layers.

Our analysis discovers intriguing patterns in Procrustes similarities across layers in Figures \ref{fig:proc_sim_txt}-\ref{fig:proc_sim_cen_img}. First, both models' textual and visual tokens exhibit an almost perfectly linear behavior across layers. These findings align with the previous research \cite{Razzhigaev2024-secretlylinear}. Second, LLaVA-NeXT's textual tokens exhibit smooth linear relationships, whereas OmniFusion's curve appears jagged. In contrast, LLaVA-NeXT's image tokens show a slight declining trend in linearity with more layers (both centered and uncentered), whereas OmniFusion's behavior is more consistent. This difference in image tokens' behavior between the models may be attributed to LLaVA-NeXT's use of a larger number of less preprocessed visual tokens, making the underlying LLM learn more non-linear relationships across the layers. In contrast, OmniFusion's more intricate adapter issues more preprocessed visual tokens that require fewer transformations in the LLM's layers.

In Figures \ref{fig:proc_sim_cen_txt} and \ref{fig:proc_sim_cen_img}, the centered Procrustes similarity scores, which exclude the residuals from consideration, reveal a surprisingly low contribution of transformer block output to the residual stream for MLLMs, significantly lower than that of purely textual LLMs, as reported in \cite{Razzhigaev2024-secretlylinear}.

Figures \ref{fig:id_txt} and \ref{fig:id_img} show that the intrinsic dimensions of textual and image token embeddings differ significantly between LLaVA-NeXT and OmniFusion. Textual dimensions drop drastically within the first few layers, stabilizing around 20, with LLaVA-NeXT and OmniFusion having slightly smaller and larger values, respectively. In contrast, OmniFusion's image token dimensions remain consistently higher across layers. Again, we hypothesize that this is due to OmniFusion's complex adapter preprocessing, which applies intricate non-linear feature representations to a fixed number of image tokens, resulting in their saturation. As expected, from Figure \ref{fig:id_joint} the joint text-visual distribution is seen to have a slightly higher intrinsic dimension than the marginal distributions.

Figure \ref{fig:aniso_txt} shows a comparable smooth decline in anisotropy for both models towards the final layers, which is only partially consistent with \cite{Razzhigaev2024-shapelearning}. The curves lack a bell-shaped profile and exhibit high anisotropy in initial layers. We believe this is yet another emergent characteristic of multimodal tokens in MLLMs. Even a more surprising situation is observed for the anisotropy of visual tokens in Figure \ref{fig:aniso_img}: LLaVA-NeXT's numerous and shallow visual embeddings follow a bell-shaped curve, whereas OmniFusion's visual representations are uniformly low with a slight decrease towards the end. This suggests that OmniFusion's visual tokens are more evenly distributed in all directions due to saturation, and its visual adapter compresses information, refining more compact concepts.

The PCA-tSNE visualization results from the first, middle, and last layers are presented in Figures \ref{fig:layer_0}-\ref{fig:layer_31}. The visualizations from other layers are available in the demo Jupyter notebook of our Python package. At first, PCA reduced data into 50 principal components, then tSNE further reduced the dimension down to 2. The obtained visualizations are consistent with the conclusions discussed above. In particular, scattering of the textual and visual embeddings is observed towards the final layers, which corresponds to a decrease in anisotropy. Also, slight, almost linear variations in the embedding streams are visually noticeable upon a close examination of visualizations across neighboring layers.

\section{Conclusion}


In this work, we presented MLLM-Microscope, a novel system for analyzing the hidden structures within Multimodal Large Language Models (MLLMs).  Our investigation of LLaVA-NeXT and OmniFusion using the ScienceQA dataset uncovered a number of important findings.

We conclude that multimodal tokens in both models exhibit highly linear trajectories in both main and residual streams across transformer layers. However, LLaVA-NeXT's image tokens display a subtle decreasing trend in linearity, whereas OmniFusion's tokens maintain consistency. Image token dimensions in OmniFusion remain consistently higher across layers compared to LLaVA-NeXT. Furthermore, OmniFusion's anisotropy remains uniformly low throughout the layers. All these observations suggest that the modality fusion strategy prior to LLM input plays a crucial role in influencing the model's internal dynamics. This insight can inform future model development and optimization efforts. Specifically, incorporating cosine-similarity-based regularization \cite{Razzhigaev2024-secretlylinear} to mitigate layer linearity may be particularly beneficial for models featuring more complex modality fusion, such as OmniFusion.

Findings and analysis obtainable by our system may provide significant insights into the transformation processes within MLLMs, revealing new emergent characteristics and guiding potential improvements in model design and training procedures. Thus, we strongly believe that MLLM-Microscope will serve as a valuable resource for researchers and practitioners aiming to optimize and advance multimodal AI systems.

\section*{Limitations}

In our experimental study, we used only a single dataset, which could limit the generalization of our findings to other available datasets.

Our analysis spanned only two multimodal large language models. This could incur inductive bias in generalizing the obtained observations to the nature of MLLMs at large.

\section*{Ethics Statement}

We follow strict ethical guidelines in our AI research, focusing on transparency and responsibility. Our work analyzes publicly available transformer models and datasets, without using human subjects or sensitive data. This ensures our findings are trustworthy and reproducible.

\section*{Acknowledgements}
This research was funded by the Science Committee of the Ministry of Science and Higher Education of the Republic of Kazakhstan (grant no. AP26197157).

\bibliographystyle{acl_natbib}
\bibliography{custom}

\appendix

\section{PCA-tSNE Visualization}
\label{sec:appendix}

\begin{figure*}[htbp]
	\centering
	\subfloat[Layer 0 PCA-tSNE\label{fig:layer_0}]{
		\includegraphics[width=0.8\textwidth]{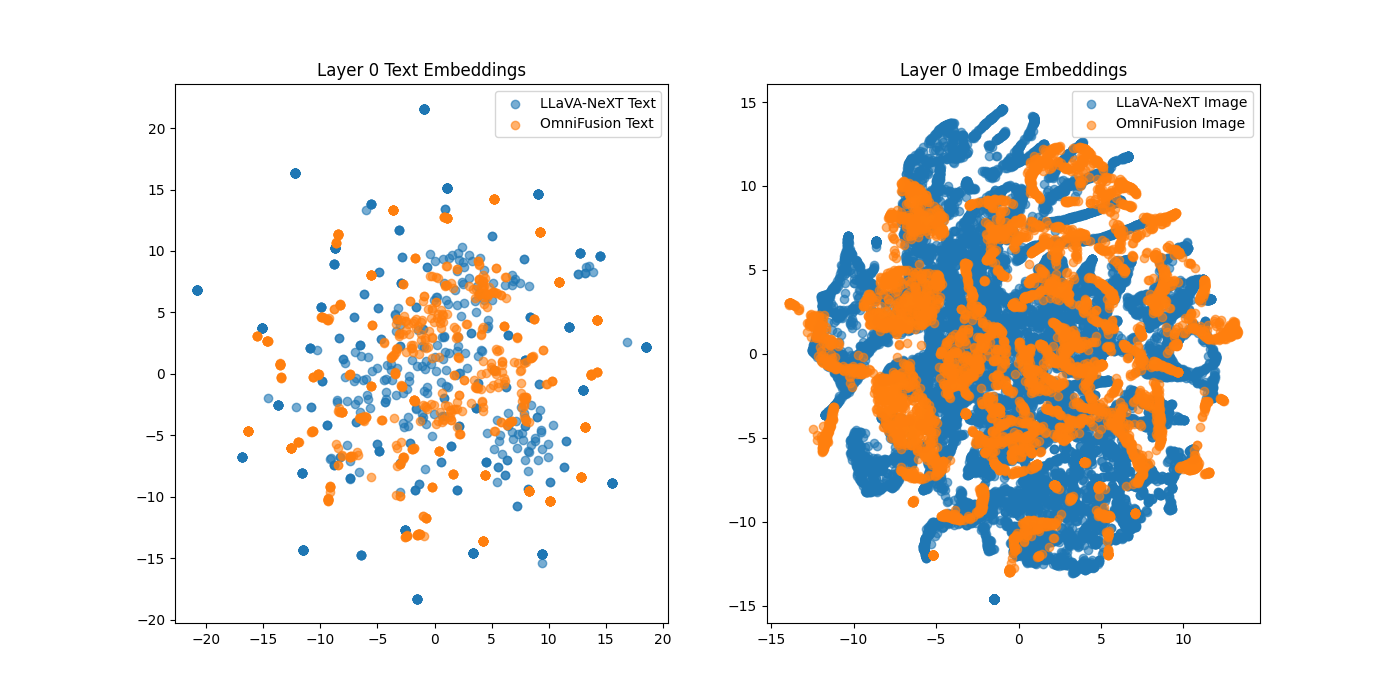}
	}
	\\
	\subfloat[Layer 15 PCA-tSNE\label{fig:layer_15}]{
		\includegraphics[width=0.8\textwidth]{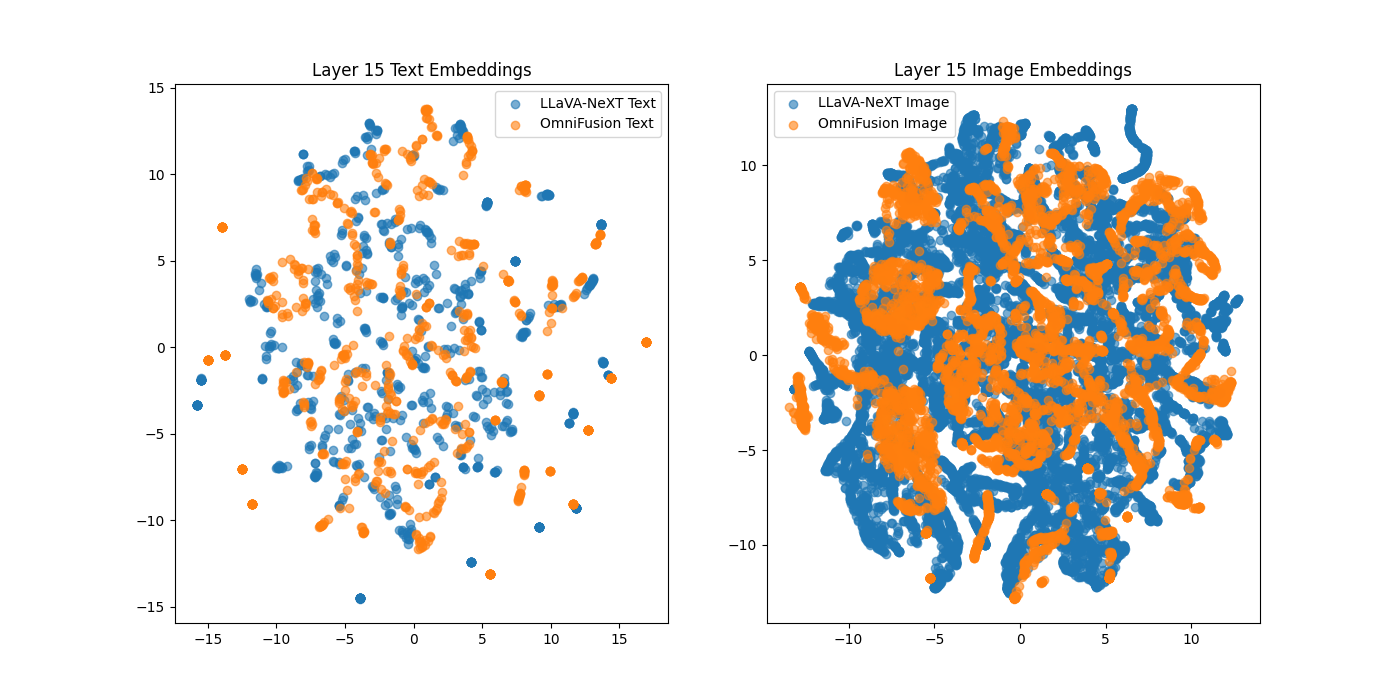}
	}
	\\
	\subfloat[Layer 31 PCA-tSNE\label{fig:layer_31}]{
		\includegraphics[width=0.8\textwidth]{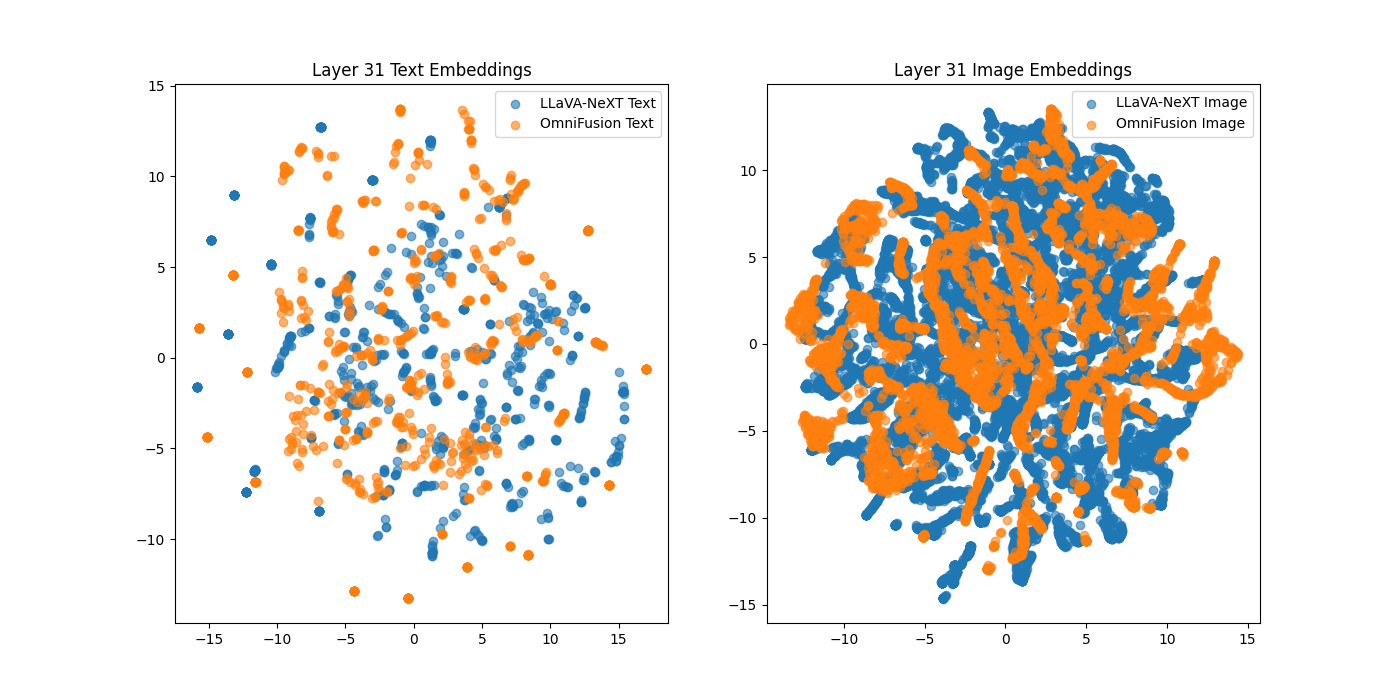}
	}
	\caption{PCA-tSNE representations of textual and visual tokens from intermediate transformer layers of the examined MLLMs.}
	\label{fig:pca_tsne_layers}
\end{figure*}

\end{document}